\documentclass[runningheads]{llncs}
\usepackage{graphicx}
\usepackage{color}
\usepackage{xcolor}
\usepackage{amsmath,amssymb,graphicx,epstopdf,cite,subfigure}
\usepackage{epsfig}
\usepackage{algorithm,algpseudocode}
\usepackage{pdfpages}
\usepackage{multirow}
\usepackage{rotating}
\usepackage{array}

\begin{document}


\title{Robust Deep Multi-modal Learning Based on Gated Information Fusion Network\thanks{This work was supported by Institute for Information \& communications Technology Promotion (IITP) grant funded by the Korea government(MSIT) (2016-0-00564, Development of Intelligent Interaction Technology Based on Context Awareness and Human Intention Understanding)}} 
\titlerunning{Robust Deep Multi-modal Learning Based on GIF Network} 


\author{Jaekyum Kim\inst{1} \and
Junho Koh\inst{1} \and 
Yecheol Kim\inst{1} \and
Jaehyung Choi\inst{2} \and \\
Youngbae Hwang\inst{3} \and 
Jun Won Choi\inst{1}\thanks{Corresponding Author}}
%

\authorrunning{Jaekyum Kim et al.} 


\institute{Hanyang University, Seoul, Korea \\
\email{\{jkkim, jhkoh, yckim\}@spa.hanyang.ac.kr, junwchoi@hanyang.ac.kr} \\ \and
Phantom AI Inc., Burlingame, CA, USA \\
\email{jaehyung@phantom.ai}\\ \and
Korea Electronics Technology Institute (KETI), Seongnam-si, Korea \\
\email{ybhwang@keti.re.kr}}

\maketitle

\begin{abstract}
The goal of multi-modal learning is to use  complimentary information on the relevant task provided by the multiple modalities to achieve reliable and robust performance. Recently, deep learning has led significant improvement in multi-modal learning by allowing for fusing high level features obtained at intermediate layers of the deep neural network. This paper addresses a problem of designing robust deep multi-modal learning architecture in the presence of the modalities degraded in quality. We introduce deep fusion architecture for object detection which processes each modality using the separate convolutional neural network (CNN) and constructs the joint feature maps by combining the intermediate features obtained by the CNNs. In order to facilitate the robustness to the degraded modalities, we employ the gated information fusion (GIF) network which weights the contribution from each modality according to the input feature maps to be fused. The combining weights are determined by applying the convolutional layers followed by the sigmoid function to the concatenated intermediate feature maps. The whole network including the CNN backbone and GIF is trained in an end-to-end fashion. Our experiments show that the proposed GIF network offers the additional architectural flexibility to achieve the robust performance in handling some degraded modalities.

\end{abstract}

\section{Introduction}

Multi-modal learning refers to a machine learning problem aiming to improve learning performance using the experience acquired from the different types of data sources. Basically, such multi-modal data delivers rich and diverse information on the phenomenon relevant to the given task.
Human is naturally born to be a good multi-modal learner in that it effectively learns from various modalities including audio, video, smell, touch, and so on. On the contrary, multi-modal fusion has been one of the most challenging problems in machine learning field due to the difficulty of combining the high level semantic information delivered by the different sources. Basically, multi-modal fusion concerns in which data processing stage the information fusion is conducted, which leads to the categorization into {\it early fusion} and  {\it late fusion} \cite{li2017semantics}. While the early fusion aims to extract the joint representation directly from the raw or preprocessed data, the late fusion aggregates the decisions separately made by the machine learning models for each modality. The late fusion is considered to be easy to implement but its performance is limited in that the correlation structure underlying in multi-modal sources is not fully utilized. Early fusion is also difficult to find a good joint representation due to significantly different data structures between modalities.  Recent emergence of deep neural network (DNN) technique (called deep learning) \cite{deeplearning} has enabled the extraction of the hierarchical semantic features from the raw data and consequently led to better and flexible use of feature-level data fusion. 
The common practice for such feature-level fusion is to construct the shared representation by merging the intermediate features obtained by separate machine learning models. In this sense, this fusion approach is referred to as {\it intermediate fusion}.
Leveraging the high level representation found by DNN, the {\it deep multi-modal learning}  (DML) technique was shown to achieve remarkable performance for a variety of multi-modal learning problems including audio-visual speech recognition \cite{Mroueh2015DeepML,noda}, multi-modal activity and emotion recognition \cite{Radu,sentanal,Kahou2015EmoNetsMD}, image analysis from RGBD data \cite{Eitel2015MultimodalDL,rgbd_rcnn,gupta2016cross}, and camera and Lidar sensor fusion \cite{mv3d,pf}.

The ultimate goal of the multi-modal learning is to achieve the highest level of reliability and robustness in performing the given task using the redundant information provided by multi-modal data.
This implies that when the information provided by a single modality is not sufficiently good enough, the multi-modal learning uses the complimentary information delivered by the different modalities and compensates for the performance degradation. 
The robustness against the degraded data quality can also be readily offered by the conventional late fusion approaches which aggregate the decisions in proportion to their credibility. On the contrary, it is not obvious how the intermediate fusion for DML can enjoy such selective information combining  since it is difficult for the machine learning models to judge the reliability of the intermediate features. One conceivable approach is to train the fusion network with the data set containing various types of degradation, hoping that the architecture learns to use only reliable features from the multi-modal sources. However, our empirical evaluation reveals that the existing fusion architectures are not flexible enough to adapt their fusion strategy  to the variation in data quality.  This quests the new DML architecture which can  take the information as needed from each modality to achieve the robust performance.

This paper proposes the DML architecture that can offer robust performance for missing or degraded modalities. Towards this end, we introduce a feature-level gated information fusion (GIF) network which combines the features obtained for each modality in a way that only information relevant to the task is aggregated. The GIF network controls the amount of information flow incoming from each modality through {\it gating mechanism}. Specifically, the GIF network selectively gates the contribution of the features by weighting each element of features by the factor between 0 and 1. These weights are independently calculated through the separate network called weight generation (WG) network. The WG network takes  all concatenated features for all modalities as an input and produces the weights by applying the convoluation layers followed by the sigmoid function.  
In fact, this operation resembles the gating operations used in long short term memory (LSTM) \cite{lstm} in that it controls the operation of information gating in a data-dependent manner. 
We build the deep 2D object detection architecture based on the proposed multi-modal fusion method. 
The proposed method first applies the multiple convolutional neural network (CNN) networks (e.g. VGG \cite{vgg}, ResNet \cite{Resnet}, etc) to generate the intermediate feature maps for the different modalities. Then, we combine these feature maps across the modalities through the proposed GIF network.  The rest of the procedure to perform the object detection based on the joint feature maps found by the GIF network follows that of the single shot detector (SSD) \cite{ssd}.  

The prior work most closely related to our work is \cite{arevalo2017gated}, in which the similar gated fusion was used to extract the joint features from the text and image data.  While the work in \cite{arevalo2017gated}  focuses on the role of gating function for modality selection, we aim to highlight the different aspect of the gated fusion for improving the robustness of deep multi-modal fusion in the context of object detection.  
The key contributions of our work are highlighted below.
\begin{itemize}
\item We demonstrate that our gated fusion network can effectively improve the robustness of  multi-modal learning. Note that developing a robust perception system using redundant sensors is a crucial problem in various safety-critical applications such as autonomous driving and mobile robot.

\item We present the robust 2D object detector built upon the proposed multi-modal fusion scheme. 
The idea of our weighted information fusion is not limited to the object detection and can readily extended to other learning models utilizing multi-modal data. 
\item In order to promote the robustness of our scheme, we train our model using the special data augmentation strategy. We generate the additional examples by corrupting some of modalities in various ways (e.g. blanking, noise addition, occlusion, severe change in illumination)  and guide our model to learn the way to fuse the different modalities with the proper weights.  
\item The experiments conducted with the SUN-RGBD dataset \cite{sunrgbd} and KITTI camera and Lidar dataset \cite{kitti} show that the proposed architecture achieves better detection accuracy than the baseline object detectors even when the subset of modalities are severely corrupted. 
\end{itemize}

The rest of the paper is organized as follows. In Section \ref{sec:related work}, we review the previous literature on the DML. In Section \ref{sec:proposed}, we present the details on the proposed GIF network and the robust 2D object detector based on multi-modal fusion. The experimental results are provided in Section \ref{sec:experiments} and the paper is concluded in Section \ref{sec:conclusion}.

\section{Related works}
\label{sec:related work}

 In this section, we briefly review the existing works on the DML methods. 

\subsection{Deep Multi-modal Learning}

The earliest research on DML goes back to the works in \cite{Ngiam} and  \cite{srivastava14b} first showing that the effective joint data representation can be found using deep models such as deep autoencoder  and deep Boltzman machine (DBM). Since then, the DML has been shown to work for a variety of learning tasks including representation learning, data fusion, translation, and alignment. (See \cite{survey} and \cite{spmagazine} for comprehensive review on DML.)  
Among them, we specifically review the multi-modal data fusion due to the relevance to our work. The multi-modal fusion aims to extract as much relevant information on the task as possible from the data having heterogeneous characteristics. 
Since the emerging DNN is good at finding high-level semantic features through the hierarchical pipelined data processing, the intermediate fusion, which combines the features found at the middle layers of the DNN, has given rise to an effective means for multi-modal fusion. 

Thus far, various DML techniques have been proposed for different types of modalities.  In  \cite{Ngiam}, the speech recognition was enhanced by using the joint data representation learned from the voice record and the video of lip movement. In \cite{Kahou2015EmoNetsMD}, the audio feature from CNN and the visual features from the deep belief network were aggregated into single video descriptor for emotion recognition. In \cite{Mroueh2015DeepML} and \cite{sentanal}, the feature-level multi-modal fusion was shown to achieve good performance in the application to speech recognition and sentiment analysis, respectively.
The DML architecture was also designed to generate the effective features for RGB-D (RGB-depth) and multi-view images.  In \cite{Eitel2015MultimodalDL}, the feature vectors obtained from the fully connected (FC) layer of two separate CNNs were combined to generate the joint features for RGB-D images.  In \cite{rgbd_rcnn}, the performance of the RGB-D fusion was improved by finding the effective encoding scheme for depth image.
In \cite{li2017semantics}, multi-level fusion architecture was proposed to learn multi-modal features for semantic segmentation. 

\subsection{Object Detection Using Multi-modal Data}


Recently, the CNN led to remarkable performance improvement for the recognition of 2D image.
Thus far, various CNN-based object detectors have been proposed. Basically, these object detectors calculate the score for the bounding box candidate and the object class based on the feature map produced by the CNN. The state-of-the-art object detectors include the faster R-CNN \cite{fasterrcnn}, SSD \cite{ssd},  YOLO \cite{yolo}, and YOLOv2 \cite{yolo2}.
The object detection can be extended for the multi-modal setup. 
In \cite{xu2017multi}, object detection based on RGB-D data was performed using the cross-modality feature found by three CNN architectures. In \cite{hoffman2016learning}, the deep fusion scheme based on RGB-D image was proposed using the {\it hallucination network} which learns a new RGB image representation by mimicking the depth network.
In \cite{mv3d}, the multi-view images are constructed from raw Lidar measurement data and used to perform 3D object detection along with RGB image in the context of automated driving. In \cite{pf}, the authors proposed the {\it point-fusion network} which predicts the corner location of the 3D bounding box based on the Lidar 3D point data.

\section{Robust deep multi-modal learning (R-DML)}
\label{sec:proposed}

In this section, we present our robust deep multi-modal learning (R-DML) architecture in details. 


\subsection {R-DML Architecture}

 \begin{figure*} [t]
  	\centering
    \centerline{\epsfig{figure=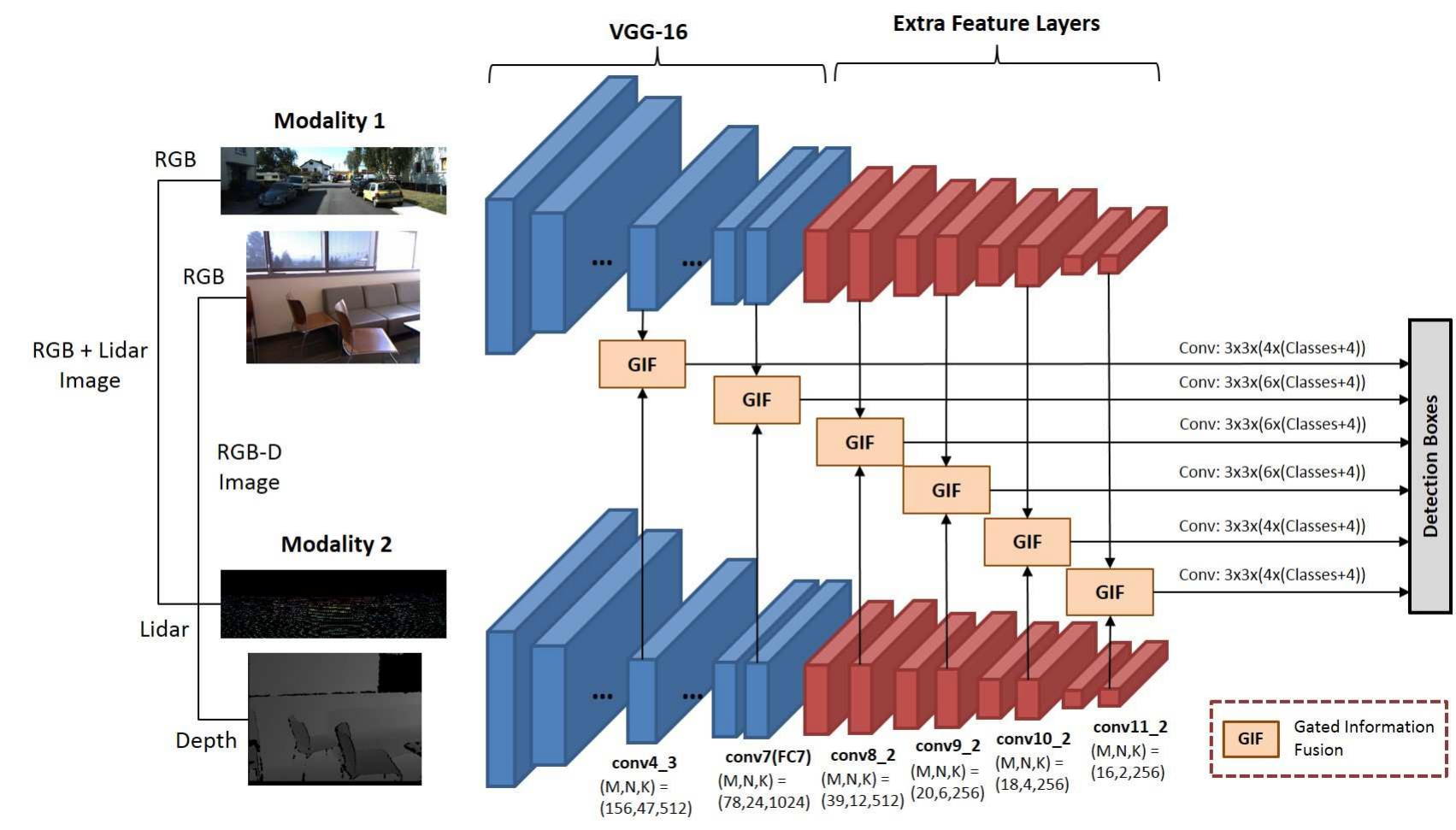, width=120mm, height=65mm}}
    \caption {Overall structure of the proposed R-DML. The R-DML takes the intermediate feature maps from both modality 1 and modality 2 using separate CNNs and combines them through the proposed GIF network. The joint feature maps produced by the GIF network are used to compute the score for object detection following the procedure of SSD.  }
    \label {R-DML}
 \end{figure*} 

\subsubsection {Overall Architecture}
The structure of the proposed R-DML is described in Fig.~\ref{R-DML}. Though our idea can be applied to the case of more than two modalities, we consider the example of two modalities.  First, two separate CNN pipelines are used to extract the intermediate features to be fused. Each CNN consists of the CNN backbone network (e.g. VGG-16) followed by 8 extra convolutional layers. This configuration is similar to that of SSD. We combine the feature maps at the layers of conv4$\_$3, conv7 (FC7), conv8$\_$2, conv9$\_$2, conv10$\_$2, and conv11$\_$2 layers.\footnote{We follow the notations of the SSD in \cite{ssd}.} These joint feature maps are used to perform object detection in different scales. As shown in Fig.~\ref{R-DML}, the GIF network is employed for feature-level information fusion. The GIF adjusts the contribution of the feature maps from each modality adaptively, whose detailed operation will be described next. In order to validate the benefit of the GIF, we compare our method with the baseline object detector referred to as the baseline DML (B-DML), which has the same structure as R-DML except that the combining weights in the GIF network are fixed to one. 

 \begin{figure}[t]
 	\centering
    \centerline{\epsfig{figure=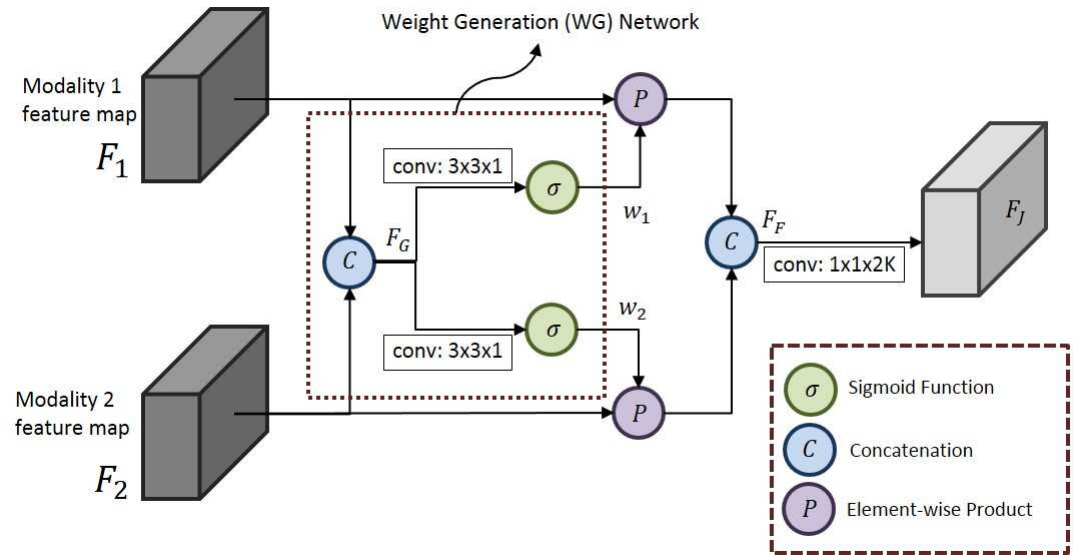, width=100mm, height=50mm}}
 	\caption {The structure of the proposed GIF network. The GIF network produces the weight maps $\mathbf{w}_1$ and $\mathbf{w}_2$ by applying the convolutional layer and sigmoid function to the input features. Then, $\mathbf{w}_1$ and $\mathbf{w}_2$ are multiplied to the feature maps $\mathbf{F}_1$ and $\mathbf{F}_2$ for weighted information fusion. }
    \label {gif}
 \end{figure}

\subsubsection {Gated Information Fusion (GIF) Network}
Fig.~\ref{gif} depicts the structure of the GIF network. The GIF network takes the intermediate feature maps from each CNN as an input and combines them with the weights calculated by the WG network. Let  $\mathbf{F}_{1}$ and $\mathbf{F}_{2}$ be the $M \times N \times K$ feature maps obtained by two CNNs corresponding to two input modalities. The actual values of $M, N$ and $K$ for each layer are provided in Fig.~\ref{gif}. The GIF network consists of two parts: 1) the information fusion  network and 2) the WG network.  The information fusion network multiplies the $M \times N $ weight maps $\mathbf{w}_1$ and $\mathbf{w}_2$ to the feature maps $\mathbf{F}_{1}$ and $\mathbf{F}_{2}$, respectively. Such multiplication is done in element-wise for each feature map. Then, the weighted feature maps are concatenated across all modalities and $1 \times 1 \times 2K$ convolution is applied to fuse the feature maps. These operations result in the joint feature maps $\mathbf{F}_J$. 
Meanwhile, the WG network calculates the weights based on the input features as shown in Fig~\ref{gif}. After concatenating the feature maps over all modalities, two separate $3 \times 3 \times 1$ CNN kernels $\mathbf{C}_1$ and $\mathbf{C}_{2}$ are applied in parallel to increase the depth in generate the high level features, which are used to calculate the combining weights \footnote{Our extensive experiments show that additional depth over single convolutional layer does not help improving the effectiveness of the gating operation.}.  Then, the sigmoid function is applied to produce the weight maps $\mathbf{w}_1$ and $\mathbf{w}_{2}$.  
We summarize the operation of the GIF network in the following equations.

\begin{align}
\mathbf{F}_{G} &= \mathbf{F}_{1} \boxplus \mathbf{F}_{2} \\
\mathbf{w}_{1} &= \sigma(\mathbf{C}_{1} * \mathbf{F}_{G} + \mathbf{b}_{1}) \\
\mathbf{w}_{2} &= \sigma(\mathbf{C}_{2} * \mathbf{F}_{G} + \mathbf{b}_{2}) \\
\mathbf{F}_{F}(i) &= (\mathbf{F}_{1}(i) \odot \mathbf{w}_{1}) \boxplus (\mathbf{F}_{2}(i) \odot \mathbf{w}_{2}), \;\;\;\; i = 1,...,K, \\
\mathbf{F}_{J} &= ReLU(\mathbf{C}_{J} * \mathbf{F}_{F} + \mathbf{b}_{F})
\end{align}
where
\begin{itemize}
    \item $\sigma({{x}}) \triangleq \frac{1}{1+e^{-{x}}}$: sigmoid function(element-wise)
    \item $x * y$: convolutional layer
	\item $x \odot y$: element-wise product
    \item $x \boxplus y$: concatenation
   \item $\mathbf{F}(i)$: $i$th feature map of $\mathbf{F}$
    \item $\mathbf{b}_{F}, \mathbf{b}_{1}, \mathbf{b}_{2}$: biases of the convolution layers.
\end{itemize}

\subsection {Training}
\subsubsection {Data Augmentation}
In order to guide our network to learn to fuse the features appropriately in adverse environments, we design the data augmentation method. We generate the new training examples by applying various types of degradation to the subset of modalities. With such diverse training examples, our model would learn the robust multi-modal fusion. In our work, various type of modifications can be applied for data augmentation, including
\begin{itemize}
\item Blank Data (Type 1): we feed all pixel value to zero. 
\item Random occlusion (Type 2): we occlude the object using the black box whose size and location are randomly chosen.
\item Severe illumination change (Type 3): we brighten the image in the rounded local region where the center and radius of the region and the brightness are randomly chosen.
\item Additive random noise (Type 4): we add the random Gaussian noise where noise variance is randomly chosen within the certain range. 
\item No action.
\end{itemize}
The type of modification and which modalily will be degraded are chosen randomly with equal probability. Note that this data augmentation strategy is critical for our method to achieve the robust performance for the scenarios where some of modalities are corrupted.

\subsubsection{Training Setup} 
 Except for our data augmentation strategy, we use the same training setup used in SSD (e.g., matching strategy, hard negative mining, and multi-task loss function). We use  VGG-16 pretrained model on ImageNet in two parallel CNN pipelines. The stochastic gradient descent (SGD) are used with the mini-batch size of $2$  and the momentum parameter of $0.9$. We set the initial learning rate to $0.0005$. We set the weight decay parameter applied to L2 regularization term to $0.0005$.

\section{Experimental results} 
\label{sec:experiments}

In this section, we evaluate the performance of the proposed R-DML scheme using two public datasets: KITTI dataset \cite{kitti} and SUN-RGBD dataset \cite{sunrgbd}. We first investigate the behavior of the gating operation to verify the effectiveness of the GIF network. Then, we compare the performance of our scheme with that of other multi-modal fusion schemes. Note that for fair comparison, we re-trained other algorithms using the same augmentation method as that used to train the R-DML.  A total of $130$ epochs and $200$ epochs are executed with the KITTI dataset and SUN-RGBD dataset, respectively.

\subsection{Datasets}
\subsubsection{KITTI Dataset}
The KITTI dataset is collected by driving the car equipped with with Pointgrey cameras and a Velodyne HDL-64E Lidar in various driving scenarios. The training set and test set contain 7,481 images and 7,518 images, respectively. Since the labels of the test images are not publicly available, we split the labeled training dataset into the training set and validation set by half as done in \cite{3dop}. We evaluate the 2D detection performance with three object categories, i.e., car, pedestrian, and cyclist and  three difficult levels, i.e., easy, moderate, hard as proposed in the KITTI Benchmark. 

We consider the multi-modal fusion task which performs object detection using both RGB image and 3D lidar data. In order to preprocess the data for our object detector, we convert the 3D point cloud data into the 2D image in camera plane.  
The 3D point data in KITTI dataset contains  the 3D coordinate $(X,Y,Z)$ and the reflectivity $R$ measured for each reflected laser pulse. 
Specifically, we map the 3D coordinate $(X,Y,Z)$ of Lidar data into the 2D coordinate $(x,y)$ on camera plane using
\begin{align}
\begin{bmatrix}x\\ y\end{bmatrix} &= calib\_matrix \cdot \begin{bmatrix}X\\ Y \\ Z\end{bmatrix}. 
\end{align}
where $calib\_matrix$ is the matrix for coordinate transformation. Note that we quantize $(x,y)$ to the nearest integer and limit the maximum range of $(x,y)$ by that of camera plane.  For the given 2D coordinate $(x,y)$, we create three channel image by encoding the values of $X$, $Z$, and $R$ to the pixel values. This creates the image with the depth, height, and intensity (DHI) channels. The pixel values for the DHI channels are obtained by
\begin{align}
val_d &= 255 \cdot (1 - \min [X/{ max\_X},1]) \\
val_h &= 255 \cdot (1 - \min [Z/{ max\_Z},1]) \\
val_i &= 255 \cdot (1 - \min [R/{ max\_R},1]). 
\end{align}
Note that $X \in [0,max\_X]$, $Z \in [0,max\_Z]$, and $R \in [0,max\_R]$ are mapped to the pixel values between $[0,255]$ in a linear scale. For example, we set $max\_X$, $max\_Z$, and $max\_R$ to $80$ meter, $6$ meter, and $0.7$. Note that the DHI Lidar image and the RGB camera image of the size $1242 \times 375$ are used as the multi-modal inputs for the proposed object detector. We apply data augmentation to these images. Since it is hard to introduce noise and illumination change to the Lidar image, we apply them only for RGB image. 

\begin{figure}[t] 
    \centering
    \begin{subfigure}[Original RGB image]
        {\epsfig{figure=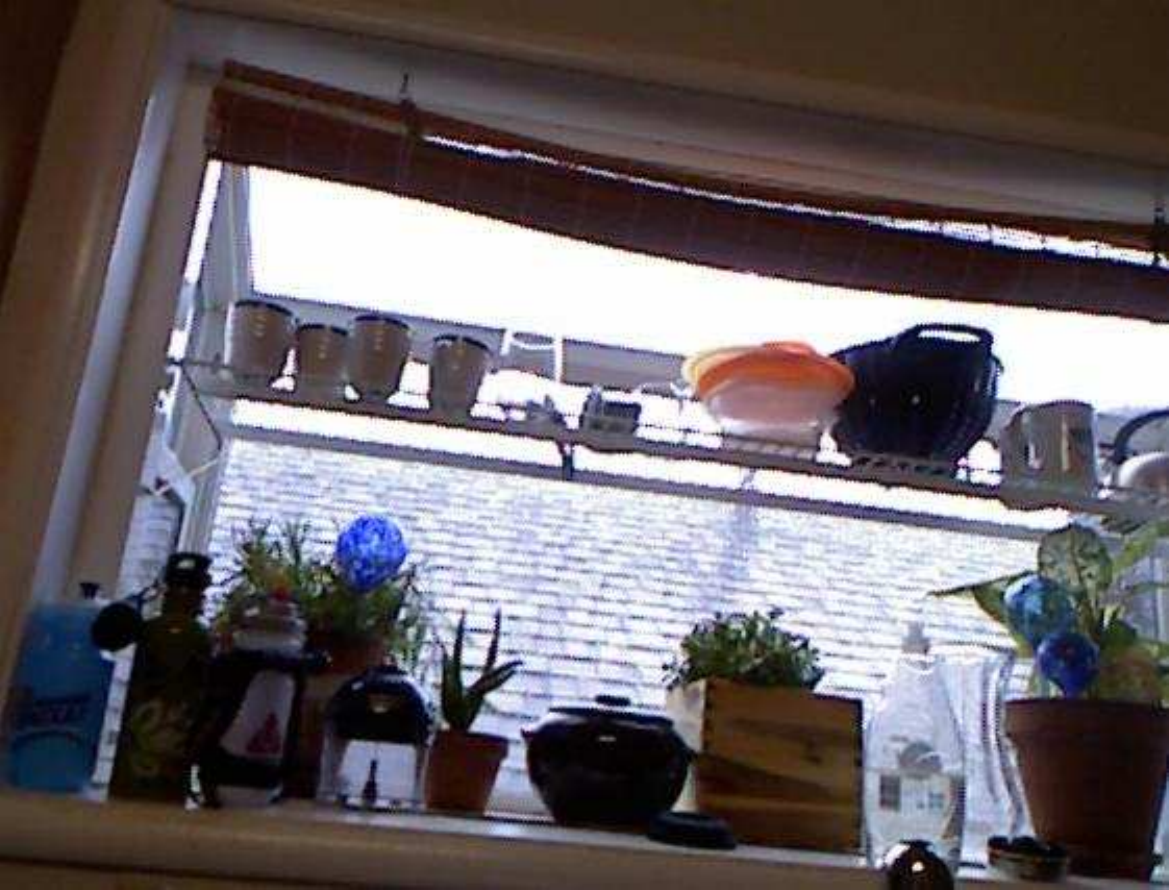, width=40mm, height=30mm}}
        \label{fig:before}
	\end{subfigure}
	\begin{subfigure}[Illumination change]
 		{\epsfig{figure=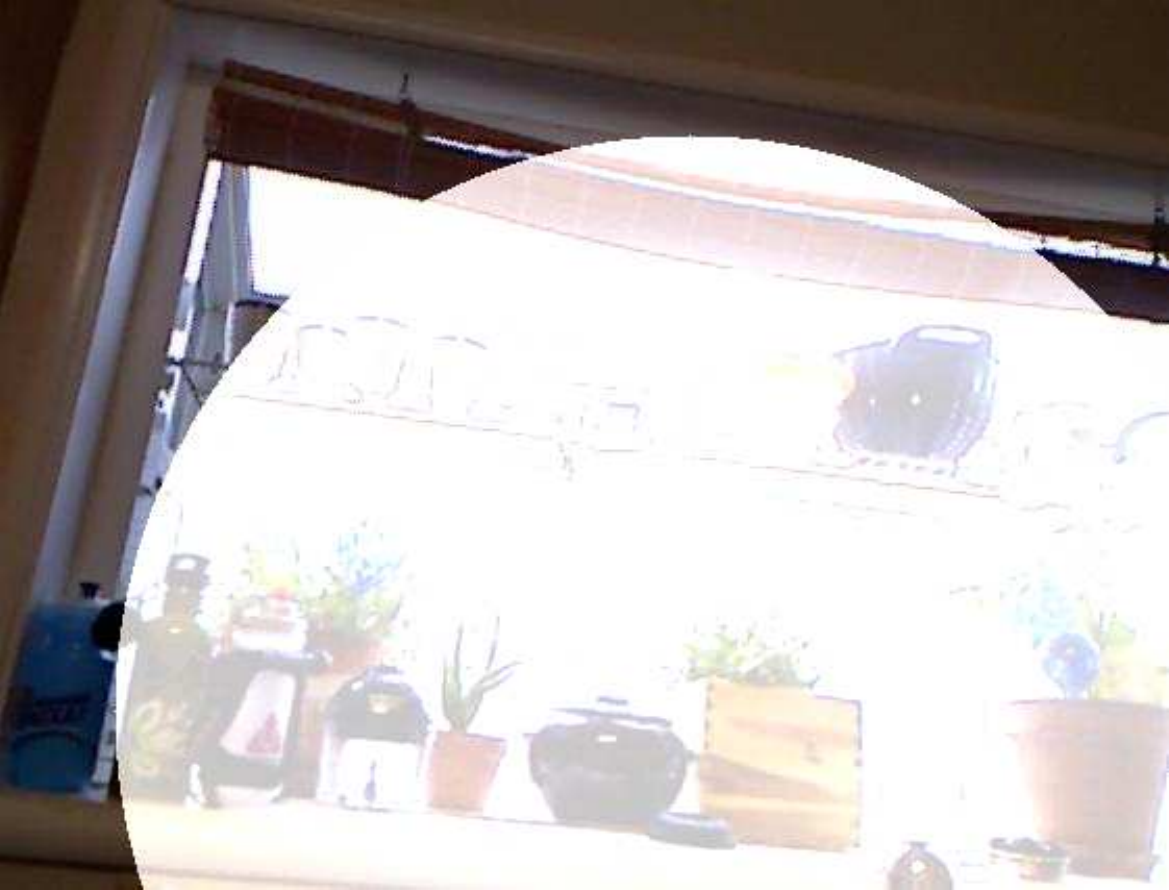, width=40mm, height=30mm}}
        \label{fig:after}
    \end{subfigure}
    \begin{subfigure}[Noise]
        {\epsfig{figure=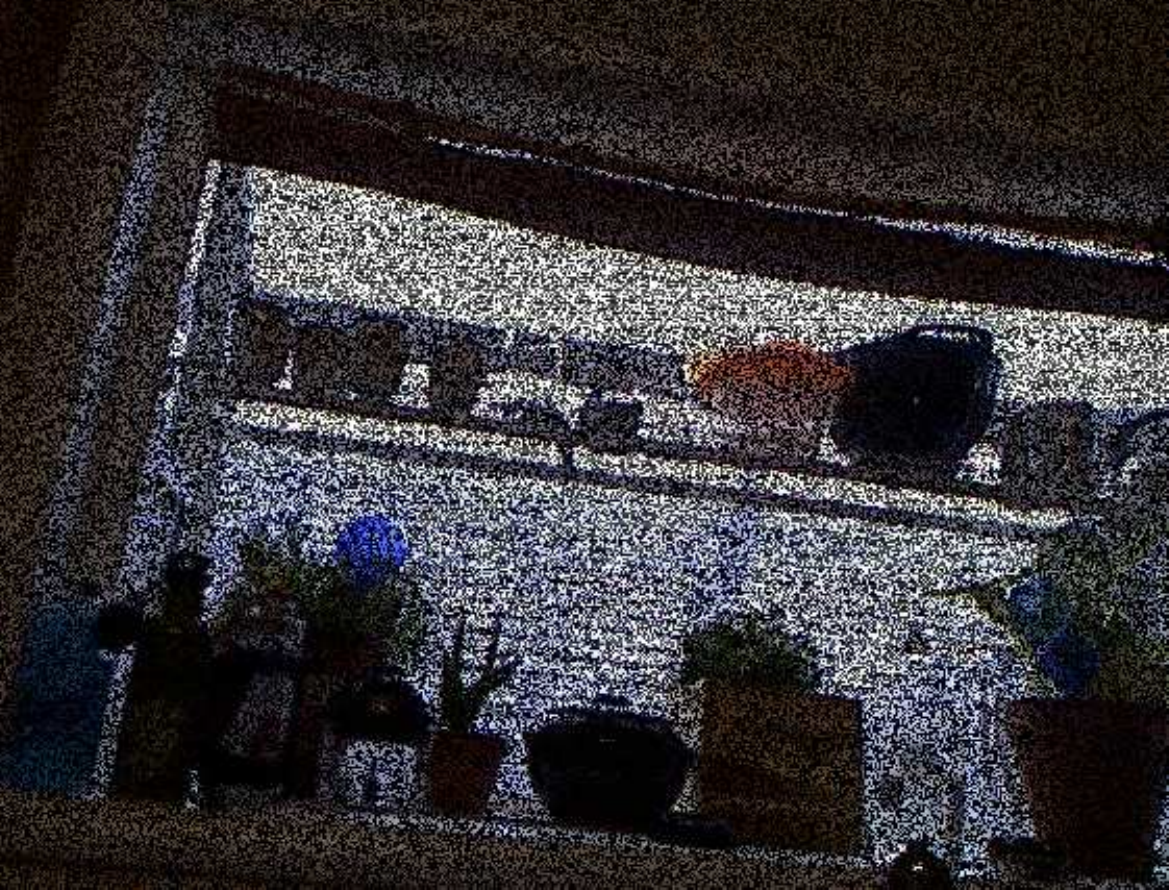, width=40mm, height=30mm}}
        \label{fig:dummy}
	\end{subfigure}
	\begin{subfigure}[Occlusion]
 		{\epsfig{figure=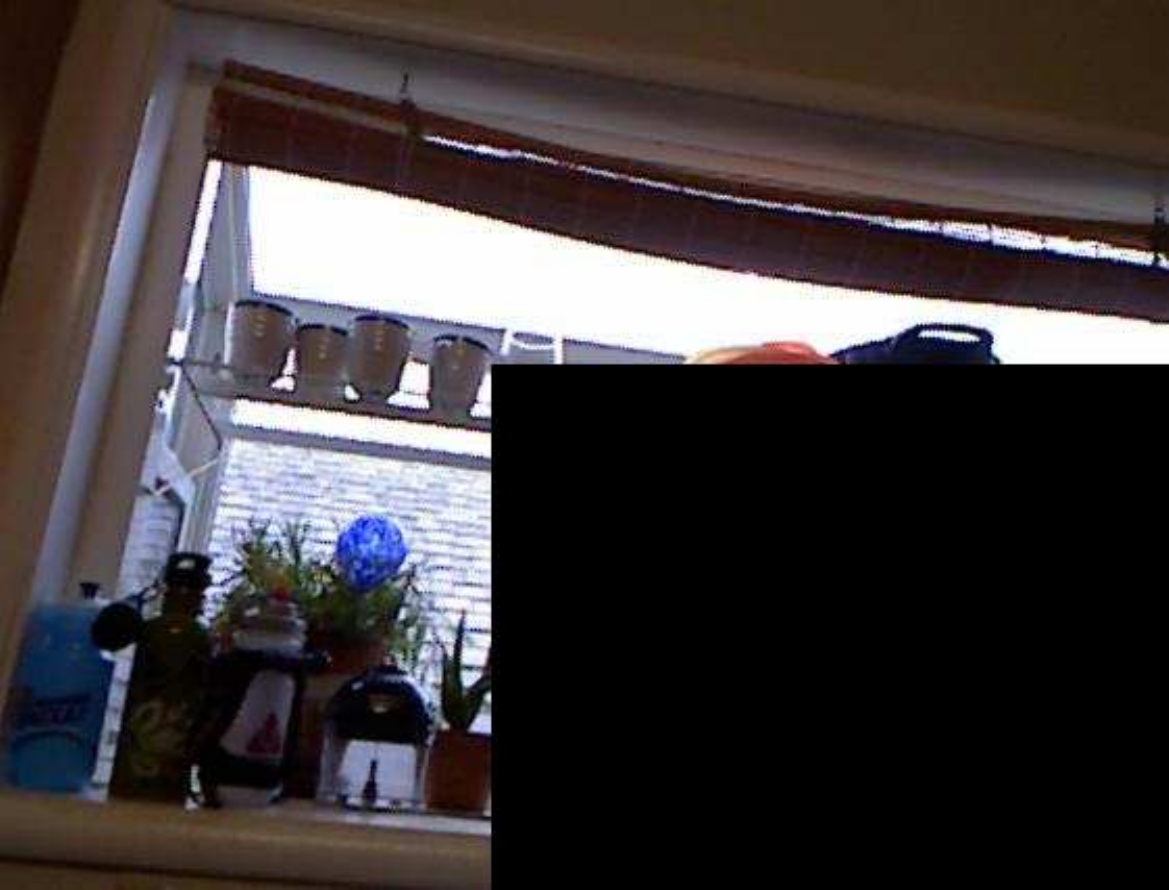, width=40mm, height=30mm}}
        \label{fig:normal}
    \end{subfigure}
    \caption{Examples of  modifications applied to the camera image on SUN-RGBD dataset.}
    \label{input_example}
 \end{figure}

\subsubsection{SUN-RGBD Dataset}
The SUN-RGBD dataset is a large-scale RGB-D dataset collected in indoor environments. It contains  10,335 RGB image and depth image pairs including NYUDv2 depth \cite{nyud2}, Berkeley B3DO \cite{b3do}, and SUN3D \cite{sun3d}. The dataset consists of 5,285 training set and 5,250 testing set. We evaluate the detection performance with 19 object categories as in \cite{sunrgbd}. Note that we apply the same data augmentation strategy used for the KITTI dataset and we set  the size of the input image to $530 \times 400$. The examples of the modifications applied to the RGB camera image in SUN-RGBD dataset are illustrated in Fig. \ref{input_example}. 

\subsubsection{Extended Test Dataset}
To evaluate the robustness of the proposed R-DML, we randomly generate the test dataset using the same types of data modification applied for the data augmentation. Both KITTI and SUN-RGBD datasets contain the RGB camera image while the modality 2 corresponds to the Lidar image and depth image, respectively. In our experiments, we come up with the following test cases: 
\begin{itemize}
\item Total: Test with all normal and degraded examples together.
\item RGB+modality2: Test with the normal test examples without any degradation.
\item RGB (blank)+modality2: Test with the test examples with RGB image blanked. 
\item RGB+modality2 (blank): Test with the test examples with modality 2 blanked.
\item RGB (occlusion)+modality2: Test with the test examples with RGB image occluded.
\item RGB+modality2 (occlusion): Test with the test examples with modality 2 occluded.
\item RGB (noise)+modality2 : Test with the test examples with the noise of the RGB image changed.
\item RGB (illumination)+modality2 : Test with the test examples with the illumination of the RGB image changed.
\end{itemize}
Note that  the performance evaluation is performed with the same number of test examples for each case.

\newcolumntype{P}[1]{>{\centering\arraybackslash}p{#1}}
\begin{table*}[t] 
\caption{Detection performance (AP) for car category on the extended KITTI test dataset}
\centering
\begin{tiny}
\begin{tabular}{|c|c|c|c|c|c|c|c|c|c|P{0.70cm}|P{0.70cm}|P{0.70cm}|c|c|c|}
\hline
\multirow{2}{*}{Test Input} & \multicolumn{3}{|c|}{Our R-DML} & \multicolumn{3}{|c|}{B-DML} & \multicolumn{3}{|c|}{Early fusion} & \multicolumn{3}{|c|}{SSD-based fusion \cite{ssd}} & \multicolumn{3}{|c|}{AVOD \cite{avod}} \\ \cline{2-16}
{} & Easy & Mod. & Hard & Easy & Mod. & Hard & Easy & Mod. & Hard & Easy & Mod. & Hard & Easy & Mod. & Hard\\ \hline
\multirow{2}{*}{Total}  & \multirow{2}{*}{\textbf{93.95}} & \multirow{2}{*}{\textbf{86.70}} & \multirow{2}{*}{\textbf{78.05}} & \multirow{2}{*}{89.86} & \multirow{2}{*}{82.21} & \multirow{2}{*}{72.21} & \multirow{2}{*}{91.10} & \multirow{2}{*}{85.65} & \multirow{2}{*}{75.83} & \multirow{2}{*}{89.69} & \multirow{2}{*}{82.03} & \multirow{2}{*}{72.96} & \multirow{2}{*}{-} & \multirow{2}{*}{-} & \multirow{2}{*}{-}\\ 
{} & {} & {} & {} & {} & {} & {} & {} & {} & {} & {} & {} & {} & {} & {} & {}\\ \hline
{RGB +}  & \multirow{2}{*}{\textbf{98.69}} & \multirow{2}{*}{\textbf{90.31}} & \multirow{2}{*}{\textbf{82.16}} & \multirow{2}{*}{93.61} & \multirow{2}{*}{87.01} & \multirow{2}{*}{77.52} & \multirow{2}{*}{95.84} & \multirow{2}{*}{89.94} & \multirow{2}{*}{79.67} & \multirow{2}{*}{91.72} & \multirow{2}{*}{87.93} & \multirow{2}{*}{78.46} & \multirow{2}{*}{89.85} & \multirow{2}{*}{87.99} & \multirow{2}{*}{80.27}\\ 
Lidar & {} & {} & {} & {} & {} & {} & {} & {} & {} & {} & {} & {} & {} & {} & {}\\ \hline
{RGB (blank)}  & \multirow{2}{*}{88.86} & \multirow{2}{*}{78.12} & \multirow{2}{*}{69.68} & \multirow{2}{*}{86.56} & \multirow{2}{*}{74.30} & \multirow{2}{*}{64.71} & \multirow{2}{*}{\textbf{89.94}} & \multirow{2}{*}{\textbf{78.99}} & \multirow{2}{*}{69.56} & \multirow{2}{*}{87.92} & \multirow{2}{*}{77.83} & \multirow{2}{*}{69.11} & \multirow{2}{*}{86.42} & \multirow{2}{*}{69.82} & \multirow{2}{*}{\textbf{69.77}}\\ 
+ Lidar & {} & {} & {} & {} & {} & {} & {} & {} & {} & {} & {} & {} & {} & {} & {}\\ \hline
{RGB + }  & \multirow{2}{*}{\textbf{97.39}} & \multirow{2}{*}{\textbf{90.29}} & \multirow{2}{*}{\textbf{81.84}} & \multirow{2}{*}{91.88} & \multirow{2}{*}{88.10} & \multirow{2}{*}{78.68} & \multirow{2}{*}{90.48} & \multirow{2}{*}{88.56} & \multirow{2}{*}{77.92} & \multirow{2}{*}{93.31} & \multirow{2}{*}{89.27} & \multirow{2}{*}{80.03} & \multirow{2}{*}{-} & \multirow{2}{*}{-} & \multirow{2}{*}{-}\\ 
Lidar (blank) & {} & {} & {} & {} & {} & {} & {} & {} & {} & {} & {} & {} & {} & {} & {}\\ \hline
{RGB (occl.) }  & \multirow{2}{*}{89.88} & \multirow{2}{*}{88.12} & \multirow{2}{*}{\textbf{79.03}} & \multirow{2}{*}{88.12} & \multirow{2}{*}{78.52} & \multirow{2}{*}{68.85} & \multirow{2}{*}{90.22} & \multirow{2}{*}{84.15} & \multirow{2}{*}{73.93} & \multirow{2}{*}{\textbf{91.78}} & \multirow{2}{*}{\textbf{88.22}} & \multirow{2}{*}{78.80} & \multirow{2}{*}{87.94} & \multirow{2}{*}{78.75} & \multirow{2}{*}{78.53}\\ 
+ Lidar  & {} & {} & {} & {} & {} & {} & {} & {} & {} & {} & {} & {} & {} & {} & {}\\ \hline
{RGB + }  & \multirow{2}{*}{\textbf{97.72}} & \multirow{2}{*}{\textbf{90.23}} & \multirow{2}{*}{\textbf{81.94}} & \multirow{2}{*}{92.75} & \multirow{2}{*}{87.10} & \multirow{2}{*}{77.67} & \multirow{2}{*}{90.53} & \multirow{2}{*}{88.91} & \multirow{2}{*}{79.07} & \multirow{2}{*}{84.80} & \multirow{2}{*}{74.88} & \multirow{2}{*}{66.33} & \multirow{2}{*}{-} & \multirow{2}{*}{-} & \multirow{2}{*}{-}\\ 
Lidar (occl.) & {} & {} & {} & {} & {} & {} & {} & {} & {} & {} & {} & {} & {} & {} & {}\\ \hline
{RGB (noise)}  & \multirow{2}{*}{89.33} & \multirow{2}{*}{80.15} & \multirow{2}{*}{71.12} & \multirow{2}{*}{86.75} & \multirow{2}{*}{75.13} & \multirow{2}{*}{65.71} & \multirow{2}{*}{\textbf{90.18}} & \multirow{2}{*}{\textbf{81.29}} & \multirow{2}{*}{72.04} & \multirow{2}{*}{88.67} & \multirow{2}{*}{76.12} & \multirow{2}{*}{67.18} & \multirow{2}{*}{88.88} & \multirow{2}{*}{79.79} & \multirow{2}{*}{\textbf{79.46}}\\ 
Lidar & {} & {} & {} & {} & {} & {} & {} & {} & {} & {} & {} & {} & {} & {} & {}\\ \hline
{RGB (illum.)}  & \multirow{2}{*}{\textbf{95.82}} & \multirow{2}{*}{\textbf{89.71}} & \multirow{2}{*}{\textbf{80.58}} & \multirow{2}{*}{89.37} & \multirow{2}{*}{85.31} & \multirow{2}{*}{75.87} & \multirow{2}{*}{90.48} & \multirow{2}{*}{88.42} & \multirow{2}{*}{78.60} & \multirow{2}{*}{89.69} & \multirow{2}{*}{79.96} & \multirow{2}{*}{70.82} & \multirow{2}{*}{88.60} & \multirow{2}{*}{79.33} & \multirow{2}{*}{79.00}\\ 
+ Lidar & {} & {} & {} & {} & {} & {} & {} & {} & {} & {} & {} & {} & {} & {} & {}\\ \hline

\end{tabular}
\end{tiny}
\label{GIF_table_kitti}
\end{table*}

\begin{table*}[t] 
\caption{Detection performance (AP) for car category on the extended KITTI test dataset with unseen types of modification}
\centering
\begin{scriptsize}
\begin{tabular}{|c|c|c|c|c|c|c|}
\hline
\multirow{2}{*}{Test Input} & \multicolumn{3}{|c|}{Our R-DML} & \multicolumn{3}{|c|}{B-DML} \\ \cline{2-7}
{} & Easy & Mod. & Hard & Easy & Mod. & Hard \\ \hline
{RGB + Lidar (Type1. blank)} & {-} & {-} & {-} & {-} & {-} & {-}\\ \hline
{RGB + Lidar (Type2. occl.)} & {\textbf{83.31}} & {\textbf{82.23}} & {\textbf{74.41}} & {80.50} & {77.37} & {68.89}\\ \hline
{RGB + Lidar (Type3. illum.)} & {\textbf{90.62}} & {\textbf{89.06}} & {\textbf{80.04}} & {89.70} & {87.22} & {78.59}\\ \hline
{RGB + Lidar (Type4. noise)} & {\textbf{83.10}} & {\textbf{73.34}} & {\textbf{65.67}} & {78.15} & {66.52} & {58.25}\\ \hline

\end{tabular}
\end{scriptsize}
\label{GIF_add_table_kitti}
\end{table*}

\begin{table}[t]
\caption{Detection performance (AP) on KITTI validation set. (*: trained by us, red text: ranked first, blue text: ranked second, green text: ranked third)}
\centering
\begin{scriptsize}
\begin{tabular}{|c|c|P{1.5cm}|P{1.5cm}|P{1.5cm}|}
\hline
Method & Data & Easy & Moderate & Hard\\ \hline
SSD* \cite{ssd} & Mono & 93.31 & 89.27 & 80.03\\ \hline 
3DOP \cite{3dop} & Stereo & 94.49 & 89.65 & \textcolor{teal}{80.97}\\ \hline
Mono3D \cite{mono3d} & Mono & \textcolor{teal}{95.75} & \textcolor{teal}{90.01} & 80.66\\ \hline
Deep Manta \cite{deepmanta} & Mono & \textcolor{blue}{97.58} & \textcolor{red}{90.89} & \textcolor{red}{82.72}\\ \hline
MV3D \cite{mv3d} & Lidar+Mono & 95.01 & 87.59 & 79.90\\ \hline 
SSD-based fusion* & Lidar+Mono & 91.72 & 87.93 & 78.46\\ \hline
B-DML* & Lidar+Mono & 93.61 & 87.01 & 77.52 \\ \hline
Our R-DML* & Lidar+Mono & \textcolor{red}{98.69} & \textcolor{blue}{90.31} & \textcolor{blue}{82.16}\\ \hline
\end{tabular}
\label{valid_table}
\end{scriptsize}
\end{table}

  \begin{figure}[t] 
 	\centering
    \begin{subfigure}
        {\epsfig{figure=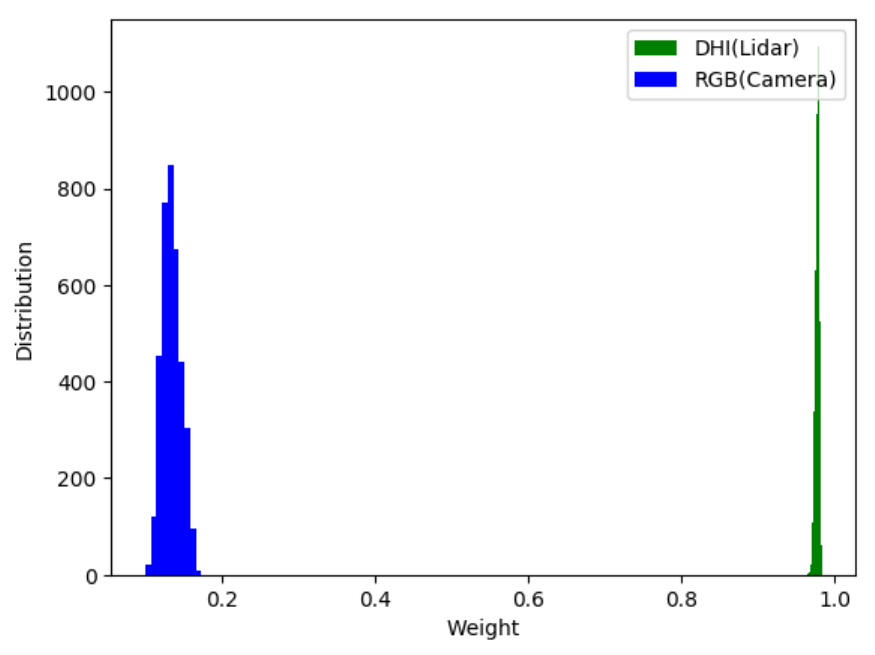, width=58mm, height=40mm}}
        \label{fig:hist_kitti}
	\end{subfigure}
 	\caption {The histogram of the averaged GIF weights at conv4\_3 layer. The weights for the blanked data are close to zero. This demonstrates the operation for reducing the contribution from unreliable data.}
    \label {hist}
 \end{figure}




  \begin{figure}[t]  
    \centering
    \begin{subfigure}[The locally occluded RGB image for test]
        {\epsfig{figure=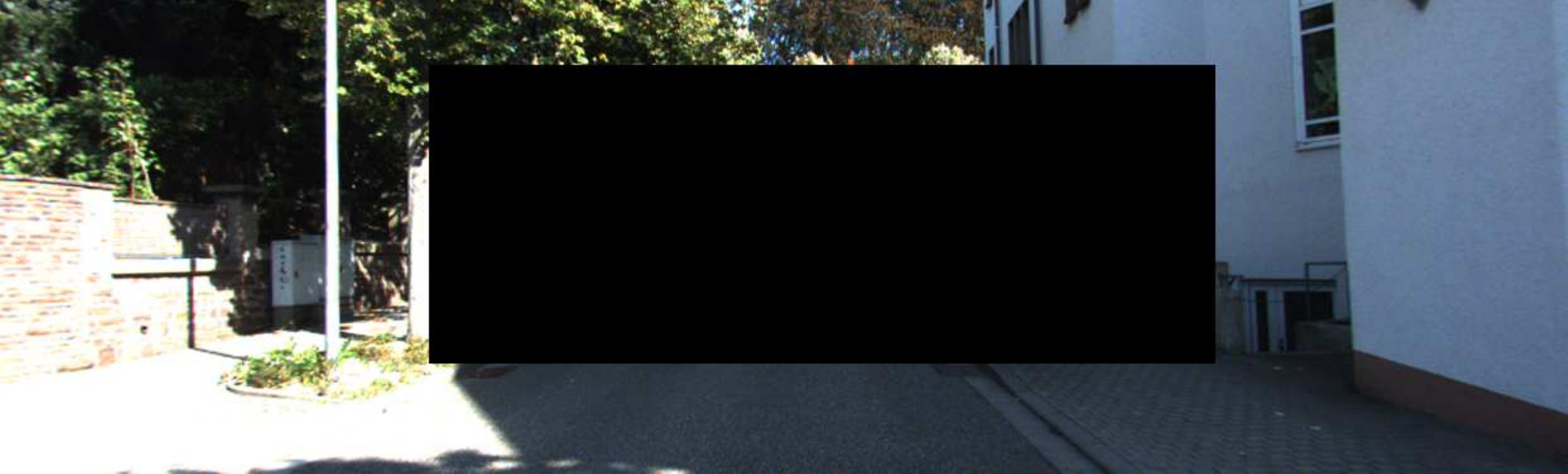, width=65mm, height=18mm}}
        \label{fig:occlusion_input}
	\end{subfigure}
    \begin{subfigure}[The weight map applied to  RGB feature maps at conv4\_3 layer]
        {\epsfig{figure=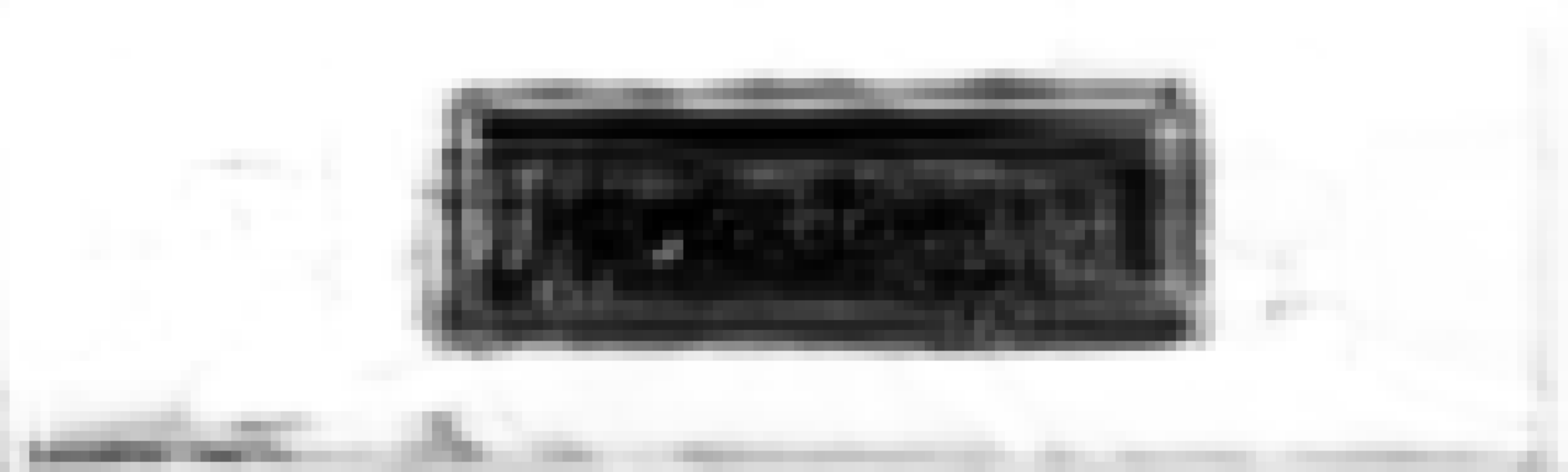, width=58mm, height=15mm}}
        \label{fig:dhi_conv4_3}
	\end{subfigure}
    \begin{subfigure}[The  weight map applied to the Lidar feature maps at conv4\_3 layer]
        {\epsfig{figure=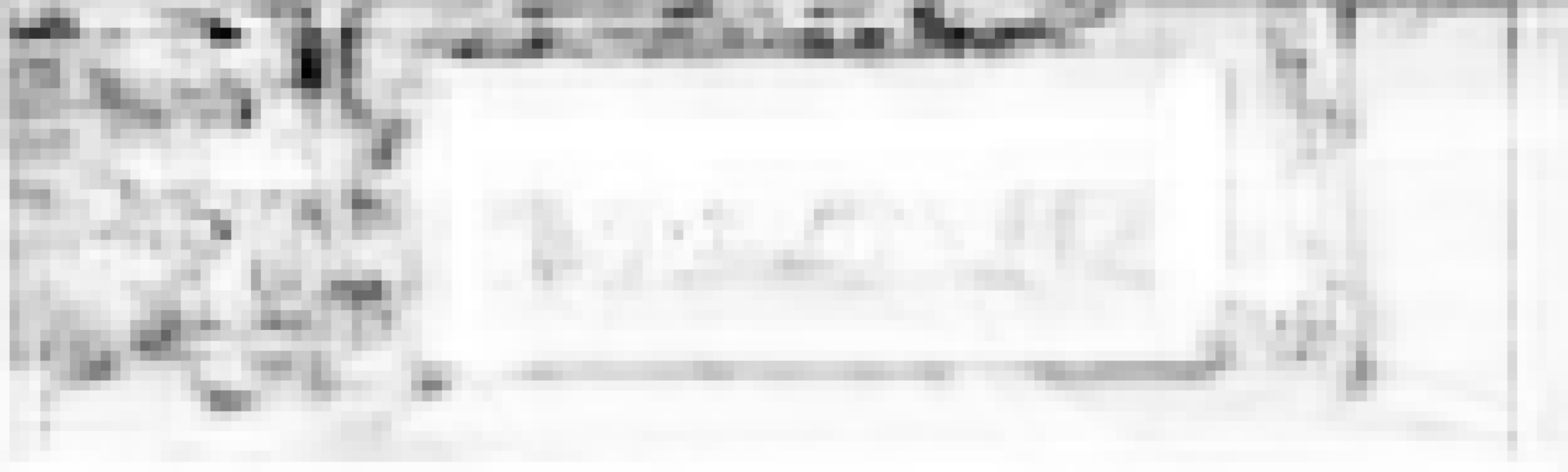, width=58mm, height=15mm}}
        \label{fig:rgb_conv4_3}
	\end{subfigure}
    \caption{The visualization of the GIF  weight maps at conv4\_3 layer. Note that the weights for the RGB features are reduced significantly for the occluded region. This shows that the gating operation conducts locally controlled information fusion.  }
    \label{weight}
 \end{figure}

 \subsection{Experimental Results on KITTI Dataset}
First, we evaluate the performance of the proposed method when tested on KITTI dataset. As a baseline algorithm, the following multi-modal fusion methods are considered:
\begin{itemize}
\item B-DML: It has the same structure with R-DML except that both gating weights applied to two modalities are fixed to one.
\item Early fusion: We concatenate two modality inputs and feed them into a single SSD.
\item SSD-based fusion: We take the late fusion approach, which combines the detection boxes generated by two SSDs. Both SSDs are  trained with the camera and Lidar images, separately. We combine the detection boxes found by two SSDs using non-maximum suppression. 
\item AVOD \cite{avod}: This is the state-of-the-art multi-modal object detection algorithm using both camera and Lidar data. Though the AVOD is capable of 3D object detection, we transform the 3D box information into the front view to compare it with our method. 

\end{itemize}
Table~\ref{GIF_table_kitti} provides the average precision (AP)  achieved by the algorithms of interest for {\it Car} category. The proposed data augmentation strategy is used for training all methods considered. The AP is evaluated using 3,740 test examples for each scenario. We observe that the proposed R-DML shows better detection accuracy than other algorithms in almost all cases. In particular, the R-DML significantly outperforms the B-DML, which shows the benefit of the proposed gated fusion method. We see that the performance gain of the R-DML over B-DML can go up to $5$\% of AP for some test scenarios (e.g. occlusion case).  Interestingly, the proposed scheme outperforms the B-DML even when the normal KITTI data is used without any data corruption for test. Since this KITTI dataset might contain some natural but somewhat benign level of real world perturbation (e.g. camera noise and adverse illumination change), this could be a part of evidence showing that the R-DML is robust to real world perturbation as well as synthetic one. In essence, all these results show that the proposed GIF network provides better model flexibility to improve the performance of multi-modal fusion. 
We evaluate the detection performance on {\it{Pedestrian}} and {\it{Cyclist}} categories as well. We obtain 70.59 (R-DML) versus 68.37 (B-DML) for moderate level for the pedestrian category and 70.11 (R-DML) versus 68.90 (B-DML) for the cyclist category. The whole results are provided in the supplemental material.

 In Table~\ref{GIF_table_kitti}, we have tested the models using the same type of modification used for training. However, the real world perturbation is hard to predict so that it is impossible to synthesize it in the training phase. Thus, the additional experiments are designed to evaluate how well the proposed method generalizes to the unseen types of modification. We train the models  using the data augmented with the type 1 to ($i-1$) modification and then test with the type $i$ modification. For example, the models trained with the dataset augmented by the type 1 (blanking) and type 2 (occlusion) modification are tested with the type 3 (illumination change) modification. 
 Table~\ref{GIF_add_table_kitti} shows that the R-DML achieves the performance gain over the B-DML when tested with each degradation type.  This shows that the proposed method exhibits the robust behavior when encountered with unseen types of degradation.

 Next, we look into the behavior of the gating operation in details.
Fig. \ref{hist} shows the histogram of the GIF weights (averaged over the whole weight map at the conv4\_3 layer) for the case that the RGB image is completely blanked. Note that the weights multiplied to the RGB features are close to zero in order to reduce the contribution from the blanked data. On the contrary, we see that the weights for the normal Lidar image are close to one. In Fig.~\ref{weight}, we visualize the GIF weight maps learned by the GIF for the case where the RGB image is locally occluded by the black box.
 We find that the GIF weights in the camera side are small only within the locally occluded region while they are high for the rest of area. Note that the GIF weights for the Lidar image are relatively high for the whole region. This shows our gating mechanism controls the amount of information combined depending on the quality of the features for each interested region.  
 
In Table~\ref{valid_table}, we compare the performance of the proposed method with other state of the art 2D object detectors when tested with the original KITTI dataset. The candidate detectors include  SSD \cite{ssd}, 3DOP \cite{3dop}, Mono3D \cite{mono3d}, Deep Manta \cite{deepmanta}, and MV3D \cite{mv3d}.  For fair comparison, we use the same train/validation split method used in \cite{3dop,mono3d,deepmanta,mv3d}. Note that SSD-based fusion, B-DML and the proposed R-DML are trained with the proposed data augmentation schemes.  We observe that the performance of the proposed object detector is better or on par with the other algorithms for all difficulty levels. This shows that the proposed fusion method exhibits competitive performance for the normal environment while promising the robust performance in the adverse environment. Note that even though the proposed R-DML is built upon the baseline SSD, significant performance gain is achieved over the baseline SSD through the multi-modal fusion strategy proposed in our work. It is interesting that the B-DML does not perform better than the SSD. This issue appears to be due to different data augmentation strategies used for training the B-DML and SSD. Due to the limitation of the SSD taking only single input modality, we could not train the SSD with our data augmentation strategy. On the other hand, the B-DML is trained with the data augmentation. We see that the B-DML does not achieve better performance than the SSD with the normal KITTI data. On the contrary when both methods are trained without data augmentation, the B-DML outperforms the SSD. Note that our R-DML significantly outperforms the B-DML and the SSD for both normal and extended KITTI datasets. 



\begin{table*} [t]
\caption{Results for detection performance (mAP) on extended SUN-RGBD test dataset}
\centering
\begin{tabular}{|c|c|c|c|c|}
\hline
{Test Input} & {Our R-DML} & {B-DML} & {Supervision transfer \cite{gupta2016cross}} \\ \hline
Total & \textbf{34.72} & 29.13 & 21.35  \\ \hline
RGB + depth & \textbf{40.43} & 36.31 & 26.68  \\ \hline
RGB (blank) + depth & \textbf{30.76} & 28.37 & 15.81 \\ \hline
RGB + depth (blank) & \textbf{32.69} & 12.03 & 22.25   \\ \hline
RGB (occlusion) + depth & \textbf{35.65} & 29.39 & 22.95   \\ \hline
RGB + depth (occlusion) & \textbf{35.04} & 33.12 & 22.75  \\ \hline
RGB (noise) + depth & \textbf{32.76} & 31.50 & 16.65 \\ \hline
RGB (illumination) + depth & \textbf{35.67} & 33.19 & 22.40  \\ \hline
\end{tabular}
\label{GIF_table_sunrgbd}
\end{table*}

\subsection{Experimental Results on SUN-RGBD Dataset} 
Table~\ref{GIF_table_sunrgbd} provides the mean average precision (mAP) of the proposed object detection algorithm.  Since there are not many recent 2D object detection algorithms using SUN-RGBD dataset, we compare our method with only the B-DML and supervision transfer (ST) methods \cite{gupta2016cross}. The ST method is the fast-RCNN \cite{fastrcnn} based object detector which combines the detection boxes obtained by two separate object detectors.  For fair comparison, we trained the B-DML and ST with the same augmentation method as that used for our R-DML.
For each test case, we use the 5,250 test examples. We see in Table~\ref{GIF_table_sunrgbd} that the proposed R-DML achieves better detection accuracy than the B-DML, which reveals the effectiveness of our gated fusion algorithm for the task of  the RGB and depth image fusion as well. Note that the R-DML maintains the performance gain over the B-DML even when the normal SUN-RGBD dataset are used for test without any modification. 
The AP results per category are provided in the supplemental material.


\section{Conclusions}
\label{sec:conclusion}

In this paper, we proposed the robust multi-modal learning technique which fuses the intermediate features produced by the CNN with appropriate contributions. Inspired by the gating mechanism used in LSTM, we devised the gated information fusion network, which combines the features from each modality with the weights computed based on the input features to be fused. Such GIF network was used to perform 2D object detection using multi-modal inputs and the whole system is trained  end-to-end. We used the special data augmentation strategy for promoting the robustness of our system, which corrupts some of modalities using various artificial operations. The experiments performed over KITTI dataset and SUR-RGBD dataset shows the superiority of the proposed method for the scenarios of missing or degraded modalities.

\bibliographystyle{splncs04}
\bibliography{bibtex/bib/egbib.bib}
\end{document}